# Method for making multi-attribute decisions in wargames by combining intuitionistic fuzzy numbers with reinforcement learning

Yuxiang Sun [1,*], Bo Yuan [2], Yufan Xue [1], Jiawei Zhou [1], Xiaoyu Zhang [1] and Xianzhong Zhou [1,*]

[1] School of Management and Engineering, Nanjing University, Nanjing 210023, China; yufanxue1@163.com (Y.X.); JiaweiZhou163@163.com (J.Z.); zhangxy0903@126.com (X.Z.)

[2] School of Electronics, Computing and Mathematics, University of Derby, Kedleston Rd, Derby DE22 1GB, UK; b.yuan@derby.ac.uk

* Correspondence: sunyuxiangsun@126.com (Y.S.); zhouxz@nju.edu.cn (X.Z.); Tel.: +86-1882-705-9351(Y.S.)

**Abstract**

Researchers are increasingly focusing on intelligent games as a hot research area.The article proposes an algorithm that combines the multi-attribute management and reinforcement learning methods, and that combined their effect on wargaming, it solves the problem of the agent's low rate of winning against specific rules and its inability to quickly converge during intelligent wargame training.At the same time, this paper studied a multi-attribute decision making and reinforcement learning algorithm in a wargame simulation environment, and obtained data on red and blue conflict.Calculate the weight of each attribute based on the intuitionistic fuzzy number weight calculations. Then determine the threat posed by each opponent's chess pieces.Using the red side reinforcement learning reward function, the AC framework is trained on the reward function, and an algorithm combining multi-attribute decision-making with reinforcement learning is obtained. A simulation experiment confirms that the algorithm of multi-attribute decision-making combined with reinforcement learning presented in this paper is significantly more intelligent than the pure reinforcement learning algorithm.By resolving the shortcomings of the agent's neural network, coupled with sparse rewards in large-map combat games, this robust algorithm effectively reduces the difficulties of convergence. It is also the first time in this field that an algorithm design for intelligent wargaming combines multi-attribute decision making with reinforcement learning.Attempt interdisciplinary cross-innovation in the academic field, like designing intelligent wargames and improving reinforcement learning algorithms.

1、**Introduction**

Artificial intelligence and machine learning are becoming more common in real-world applications, and games are increasingly fighting against humans through training agents. AlphaGo, an artificial intelligence that has achieved success in the field of Go, and Alphastar, an artificial intelligence that has achieved success in the man-machine conflict of 'StarCraft' are two typical examples. [1-2]. In RTS games, artificial intelligence methods are increasingly being integrated. In the King Glory Game, Ye D used his improved PPO algorithm to train the hero AI, with positive results. [3]. By using reinforcement learning algorithms, Silver D developed a training framework that requires no human knowledge other than the rules of the game, allowing AlphaGo to train itself, and achieving high levels of intelligence in the process [4]. Using deep reinforcement learning and supervised strategy learning, Barrigan improves the AI performance of RTS games, and defeats the built-in AI [5]. AI has become a hot research topic in recent years, showing a wide variety of applications such as deduction and analysis [6-7]. There are still insufficient effective solutions to the problem of convergence and convergence rate under a variety of conditions, especially when it comes to confrontation.

Indexes measure the value of things or the parameter of an evaluation system. It is the scale of the effectiveness of things to the subject. As an attribute value, it provides the subjective consciousness or the objective facts expressed in numbers or words. It is important to select a scientifically valid target threat assessment (TA) index and evaluate that index scientifically[8]. Target threat assessment contributes to intelligence wargame decision-making as part of current intelligent wargames. It is mainly based on rules, decision trees, reinforcement learning, and other technologies in the current mainstream game intelligent decision-making field, but rarely incorporates multi-attribute decision-making theory and methods of management into the intelligent decision-making field. The actual wargame data obtained through wargame environments is presented in this paper, as well as the multi-attribute threat assessment indicators that are effectively transformed and presented as a unified expression. Using the three expression forms of real number, interval number, and intuitionistic fuzzy number, the multi-attribute decision-making theory and method are used to analyze the target threat degree, and then the reward function in reinforcement learning is established to train more effective intelligent decision-making algorithm.For the first time, this method combines the multi-attribute decision-making of management science and reinforcement learning of control science. This method appears to be effective based on a wargame experiment.

To this end, this article conducts the following research:

（1）In this article we attempt to combine the multi-attribute decision-making method in management with reinforcement learning in cybernetics for the first time by extracting data from the environment in wargames and combining real numbers, interval numbers, and intuitionistic fuzzy numbers for data expression.Perform the multi-attribute decision analysis on this basis, analyze the threat of the opponent, and then feed back the reward function. The reward function is trained in an algorithm of reinforcement learning to obtain a more ideal result.

（2） Increase the training convergence speed. To solve the sparsity problem involved in agent training, one that leads to the emergence of agent strategies that are non-convergent or slowly convergent. The main solution is to use multi-attribute decision making（MADM）to figure out the threat of the opponent in advance and determine it in real time in a certain step, then attack based on the threat in order to obtain additional rewards and add it to the training process.

（3）Increase the winning rate of chess piece agent training.The intelligence of the agent improves by predicting the opponent's most threatening chess piece and beating it as a result of practice, which is directly reflected in the agent's winning rate against a regular opponent.

## 2、Related theories

### 2.1 The reinforcement learning

Reinforcement learning is a large category of machine learning that is based on solving interaction problems using Bellman equations [9], helping to improve and ultimately achieve the goal.Ultimately, reinforcement learning causes the agent to form a strategy that maximizes the reward value in order to achieve the goal [10]. During the 1990s, Littman proposed multi-agent reinforcement learning based on the MDP as the framework and applied the ideas and algorithms of reinforcement learning to multi-agent systems, frequently considering both competition between agents and cooperation between them [11]. Reinforcement learning starts with a Markov Process (MDP), which describes the interaction process between the agent and the environment through state and action models. As a general rule, MDP is a quadruple of $<S, A, R, T>$ consisting of 4 elements:

(1) S is a finite State Space, which represents all the states of the Agent in the environment;

(2) A is a limited Action Space, which contains the actions that the Agent is capable of taking

in each state;

(3) $R_{ss'}^{a}$ means that the agent performs a action in the s state, and the agent is rewarded by the environment interaction;

(4) The state transition function (STF) of an environment is $P_{ss'}^{a} = \mathbb{P}[S_{t+1} = s' | S_t = s, A_t = a]$, which represents the probability of performing action a on state s and transitioning to state s'.

In MDP, the agent interacts with the environment in the way shown in Figure 1.

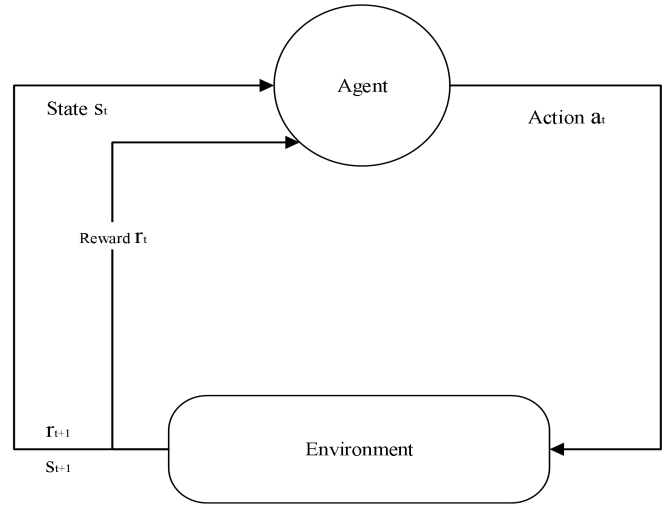

Figure 1 Schematic diagram of reinforcement learning and environment

It interprets the current environment state $s_t$ and selects $a_t$ action from its action space A; subsequently, the environment sends the corresponding reward $r_{t+1}$ to the agent, and transfers it to the new environment state $s_{t+1}$. Wait for the agent to make the next new decision [12]. When an agent interacts with its environment, there are two uncertainties: one is what kind of action to choose in state S. The strategy π (a|s) is used to represent a certain strategy of the agent (i.e. the probability distribution from state to action). And the other is the probability $P_{ss'}^{a}$ of a state transition generated by the environment. The goal of reinforcement learning is to find an optimal strategy π (a|s) so that it can obtain the maximum long-term cumulative reward in any state s and any time step t.

$$\pi^* = \operatorname{argmax}_{\pi} \mathbb{E}_{\pi} \left\{ \sum_{k=0}^{\infty} \gamma^k r_{t+k} \mid s_t = s \right\}$$

Among these, $\mathbb{E}_{\pi}$ represents the expected value given the strategy, $\gamma \in [0, 1)$ the discount rate, k the next time period, and $r_{k+t}$ the instant reward the agent receives in the time period (t + k).

In reinforcement learning, we mostly learn the optimal strategy $\pi^*$ by finding the optimal state value function $V^*(s)$ or the optimal state action value function $Q^*(s, a)$. Here are the formulas for $V^*(s)$ and $Q^*(s, a)$.

$$V^*(s) = \max_\pi \mathbb{E}_\pi \left\{ \sum_{k=0}^{\infty} \gamma^k r_{t+k} \mid s_t = s \right\}$$

$$Q^*(s,a) = \max_\pi \mathbb{E}_\pi \left\{ \sum_{k=0}^{\infty} \gamma^k r_{t+k} \mid s_t = s, a_t = a \right\}$$

Reinforcement learning has attracted researchers' attention in the field of RTS games in the recent years. In this paper, we use reinforcement learning algorithms to make intelligent decision about wargame, establish an intelligent environment for playing wargames based on reinforcement learning algorithms, use reinforcement learning algorithms to select chess pieces, and create a strong AI for game intelligence.

## 2.2 Making multi-attribute decisions

The field of decision-making has been a research hotspot in management, economics, and information science for a long time, and research on decision-making theory and methods has also been fruitful. Multiple attribute decision-making is a limited option selection problem based on multiple attributes that has a wide practical background [13][14][15]. For the multi-attribute utility theory, the value function is used to express the preference theory based on the concept of ordinal comparison and preference strength, and the utility function is used to express the preference theory based on the concept of risk choice [16]. Multi-attribute decision making (MADM) is commonly known as finite-scheme multi-objective decision-making, and it is an essential component of decision-making theory and method research [17]. The first problem of multi-attribute decision making is to determine the scheme set and the attribute set. Let A = {$A_1$, $A_2$,..., $A_n$} be the scheme set and G = {$G_1$, $G_2$, $G_3$,..., $G_M$} be the attribute set of multi-attribute decision making. The attribute value of scheme AI to attribute $G_j$ is $Y_{ij}$ (i = 1,2,..., n, j = 1,2,..., m). The decision matrix $Y_{(n \times m)}$ is composed of $Y_{ij}$. The plan set is the object of decision making. The decision matrix provides information needed to build the decision plan. Different types of analysis use the decision matrix to make decisions.

With a system with multiple targets, one target may be considered a decision-making plan, a plan set consisting of all the opponent's targets, such as tanks, helicopters, and infantry. The decision criterion is the threat level our opponent poses to our defending target. It includes a number of attributes that affect the extent of threat, such as target type, target distance, target speed, target attack ability, target defense ability, target environment, and visibility. Based on the actual wargame data obtained, apply reasonable methods to evaluate target threats.

## 2.3 The intuitionistic fuzzy number

While the application of fuzzy number theory to information processing technology has gradually matured, its limitations have slowly emerged[18]. The fuzzy number A has a unique real number in [0,1] corresponding to each of its elements. However, fuzzy number theory cannot be applied to problems of either one or the other nature at the same time. Therefore, scholars have developed fuzzy number theory. In 1986, Atanassov devised intuitionistic fuzzy numbers and explored the nature and theorems of their operations.

The intuitionistic fuzzy number definition:

In the universe of $0 \leq \pi_A(x_i) \leq 1$, the intuitionistic fuzzy set A on X has the following form:

$$A = \{\langle x_i, \mu_A(x_i), v_A(x_i) \rangle | x_i \in X\}$$

In the formula, $\mu_A(x_i): X \to [0,1]$ and $v_A(x_i): X \to [0,1]$ are the membership function and non-membership function of A, respectively, and they hold for all $x_i \in X, 0 \leq \mu_A(x_i) + v_A(x_i) \leq 1$ on A. $\pi_A(x_i) = 1 - \mu_A(x_i) - v_A(x_i)$ is the hesitation in A, and it is a measure of the hesitation of $x_i$ with A. Clearly, $0 \leq \pi_A(x_i) \leq 1$.

As a convenience, call $\alpha = (\mu_\alpha, v_\alpha)$ the Intuitionistic Fuzzy Number (IFN), There is $\mu_\alpha \in [0,1]$, $v_\alpha \in [0,1]$, and $\mu_\alpha + v_\alpha \leq 1$.

Definition 2: The algorithm of intuitionistic fuzzy numbers, set any intuitionistic fuzzy number to $\alpha = (\mu_\alpha, v_\alpha), \beta = (\mu_\beta, v_\beta)$, then

1) $\alpha \oplus \beta = (\mu_\alpha + \mu_\beta - \mu_\alpha \mu_\beta, v_\alpha v_\beta)$;

2) $\alpha \otimes \beta = (\mu_\alpha \mu_\beta, v_\alpha + v_\beta - v_\alpha v_\beta)$;

3） $\lambda\alpha = \left(1-(1-\mu_\alpha)^\lambda, (v_\alpha)^\lambda\right), \lambda > 0$ ;

4） $(\alpha)^\lambda = \left((\mu_\alpha)^\lambda, 1-(1-v_\alpha)^\lambda\right), \lambda > 0$ ;

Definition 3: In this calculation we assume $a_i = \langle \mu_{a_i}, v_{a_i} \rangle, i = 1, 2, \cdots, n$ is an IFN, $\mathbf{W} = (w_1, w_2, \cdots, w_n)$ is a weighted vector, $\sum_{j=1}^{n} w_j = 1, w_j \in [0,1], j = 1, 2, \cdots, n$ intuitionistic fuzzy weighted chess pieces satisfy IFWA, and it's a $\Theta^n \to \Theta$ mapping:

$$\text{IFWA}(a_1, a_2, \cdots, a_n) = w_1 a_1 \oplus w_2 a_2 \oplus \cdots \oplus w_n a_n = \left\langle 1 - \prod_{j=1}^{n}(1-\mu_{a_j})^{w_j}, \prod_{j=1}^{n}(v_{a_j})^{w_j} \right\rangle$$

Definition 4: $\tilde{a} = [a^L, a^U] = \{x \mid a^L \leq x \leq a^U \, a^L, a^U \in \mathbf{R}\}$ is called an interval number, $a^L$ is the lower limit of the interval, and $a^U$ is the upper limit. When $a^L = a^U$, $\tilde{a}$ degenerate into a real number, the interval number can be regarded as an extension of the real number. Here are the interval numbers $\tilde{a} = [a^L, a^U]$ and $\tilde{b} = [b^L, b^U]$, and the real number $\lambda \geq 0$. The interval number arithmetic rules are as follows:

1） Addition: $\tilde{a} + \tilde{b} = [a^L + b^L, \; a^U + b^U]$

2） Subtraction: $\tilde{a} - \tilde{b} = [a^L - b^U, a^U - b^L]$

3） Number multiplication: $\lambda \tilde{a} = [\lambda a^L, \lambda a^U]$, particularly when $\lambda = 0$, $\lambda \tilde{a} = 0$;

4） Multiplication:
$$\tilde{a} \cdot \tilde{b} = \left[\min\{a^L b^L, a^L b^U, a^U b^L, a^U b^U\} \max\{a^L b^L, a^L b^U, a^U b^L, a^U b^U\}\right];$$

5） Division: $\tilde{a}/b = [a^L, a^U] \cdot [1/b^U, 1/b^L], b^L, b^U \neq 0$

### 3、Wargaming multiple attribute index threat quantification

Obtaining scientific evaluation results requires a reasonable quantification of indicators. An important aspect of decision-making assistance in wargames is target threat assessment, and the evaluation result directly affects the effectiveness of wargame AI [19]. The aim of this section is to introduce threat quantification methods for different types of indicators. By combining the target type, this section divides the target into target distance threat, target attack threat, target speed threat, terrain visibility threat, environmental indicator threat, and target defense value. The acquired confrontation data are incorporated into different indicator types, and then the corresponding comprehensive threat value is calculated. In the table 1 are the attributes and meanings of specific indicators.

Table 1 A list of indicator attributes and their meanings

| Indicator | Attribute | Meaning |
|---|---|---|
| Target distance threat | Cost type | Distance between the two parties will influence the kill probability |
| Target attack threat | Benefit type | Threat degrees should be determined by the opponent's type, range, and lethality of the weapon |
| Target speed threat | Benefit type | The threat of speed from our opponents |
| Terrain visibility threat | Intervisibility > no intervisibility | Whether or not the terrain is visible will directly impact the threat |
| Environmental indicator threat | Benefit type | While the opponent's environment is conducive to concealment, mobility is more dangerous. |
| Target defense value | Cost type | The stronger the opponent's armor, the harder it is to destroy it |

### 3.1 Quantification of threat distance indicators

Target distance is an important parameter to evaluate the threat degree of the target [20]. In the wargame environment, the distance between entities can be calculated using the ranged tool. To simplify the process, real numbers are used to determine the distance between entities. The traditional threat quantification method uses only the target distance, whereas this article considers the distance from the control point in the wargame environment. The player who reaches the control point first wins, regardless of who reaches it first. This consideration will also be a factor in the target distance indicator.

For a red tank i and a blue tank j in wargame, calculate the target distance j threat to the red tank i. The coordinates of the control point are $O(x,y)$, and $\tau_{common}$ is the strength of the tank piece needed to pass over a grid of ordinary terrain, $\tau(x,y)$ is the stamina consumed by the vehicle through special terrain, $D(J,O)$ is the distance from tank $D_{max}$ to the control point, E is the maximum distance from the imaginary boundary, and $D_{ij}$ is the grid distance between tanks i and j. Based on a comprehensive analysis of the terrain, it can be seen that for the red tank i, the closer the blue tank j is to the control point and the closer it is to the red tank, the greater the threat of the distance indicator.

Using the following formula, we can calculate the target distance threat index of red tank i and blue tank j. The threat quantification of the target distance between the blue tank j and the red tank i is designated as $\psi_i(x,y)$, the control threat value is designated as $\varphi_j(x,y)$, and the comprehensive target distance quantitative value of the blue tank j as compared to the red tank i is designated as $\phi_{ij}(x,y)$, where i and j are the number of tanks on the red side and blue side respectively.

$$\varphi_j(x,y) = \frac{\tau_{common}}{\tau(x,y)}\left(1 - \frac{D(J,O)}{D_{max}}\right)$$

$$\psi_i(x,y) = D_{max} - D_{ij}$$

$$\phi_{ij}(x,y) = \frac{1}{2}\left(\varphi_j + \psi_i\right)$$

### 3.2 Quantification of threat indicators for target speeds

Threat levels of the target speed are a function of target motion state. The faster the target moves, the faster its position and environment change, and the harder it is for us to hit it, so the greater the threat level [21]. Therefore, the target speed threat degree can be calculated based on the benefit index, that is, the greater the target speed, the greater the threat. As a reference for the target combat intention estimation, the speed direction information can be used. Here, only the scalar of the target speed is considered. As for different types of targets, their speed determines their quantified value of threat. For example, composite armored tanks, heavy tanks, and light tanks all have different speeds. For the composite armor target, $V_{composite-max}$, the heavy tank target $V_{heavy-max}$, the light tank target $V_{light-max}$, and the relative speed between the opponent's tank target $T_j$, and our evaluation node $W_i$, $V_{ij}$, quantitatively according to the target type, and the speed threat degree $T_{vij}$ as

$$T_{vij} = \begin{cases} \beta_1 \dfrac{v_{ij}}{V_{composite-max}}, & \text{Composite armored tanks} \\ \beta_2 \dfrac{v_{ij}}{V_{heavy-max}}, & \text{Heavy tanks} \\ \beta_3 \dfrac{v_{ij}}{V_{light-max}}, & \text{Light tanks} \end{cases}$$

As part of this formula, $\beta_1, \beta_2, \beta_3 \in [0,1]$ is the threat factor for composite armored tanks, heavy tanks, and light tanks, which represents the speed threat characteristics for different types of targets, respectively. The speed threats of general composite armored tanks and light tanks are greater than those of heavy tanks. The value of $\beta_1, \beta_2, \beta_3$ can be calculated in advance by professionals according to the game characteristics of a particular target, in order to effectively characterize the target's speed characteristics.

### 3.3 Quantification of threat indicators for target attacks

In calculating the attack capability of a target tank, the attack capability threat function is mainly considered:

$$C = \left[ \ln B + \ln\left(\sum A_1 + 1\right) + \ln\left(\sum A_2\right) \right] \varepsilon_1 \varepsilon_2 \varepsilon_3 \varepsilon_4$$

Assuming B is the maneuverability of the tank pawn; $A_1$ is the weapon attack capability of the tank pawn; $A_2$ is the detection capability of the tank pawn; By calculating the threat value of the target tank chess piece based on its attack capability threat function, $\varepsilon_1$ is the forward firing capability, $\varepsilon_2$ is the bomb-carrying capability, $\varepsilon_3$ is the electronic countermeasure capability, and $\varepsilon_4$ is the missile offensive capability of the tank pawn.

### 3.4 Quantification of terrain visibility threats

In the assessment of threats, whether the targets can see each other is a significant factor. In particular, tanks, which are direct-pointing weapons, are important. To aim and track a target, one must see the target. A simplified view of the blue tank targets $T_j$ and $W_i$ is shown in the figure 2.

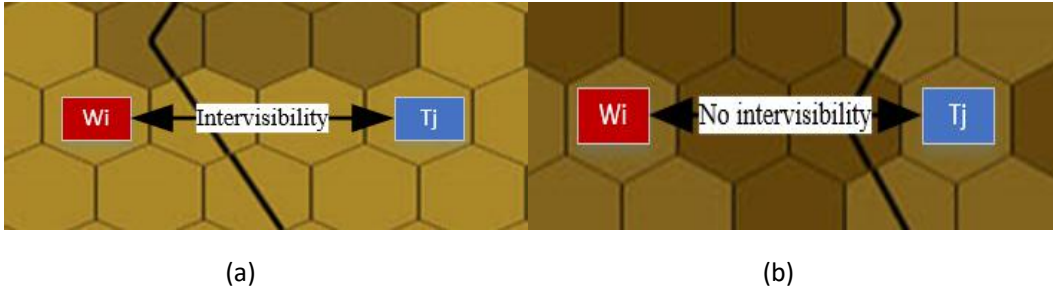

(a)             (b)

Figure 2 The visibility of the red and blue tanks.(a) Both blue and red have the same terrain elevation, and both are visible. (b) Because the terrain between the red and blue is high and the elevation on both sides is low, the red and blue will not be visible

Determine whether both parties' location information can be connected through the visibility interface of the wargame environment. A visibility interface uses the elevation between the direct connections between the two parties as a reference standard. If the direct elevation between the two parties is higher than the location of the two parties, inter-view is not possible; otherwise, it is possible. Based on the visibility interface and the locations of both parties, evaluate the terrain visibility threat $f_{ij}$ of the target blue side $T_j$ and red side $W_i$.

$$f_{ij} = \begin{cases} [t_2,1], & \max_{l \in [0,L_{ij}]} H_{ij}(l) - H_j \leq 0 \text{ 且 } \max_{l \in [0,L_{ij}]} H_{ij}(l) - H_i \leq 0 \\ [0,0], & \max_{l \in [0,L_{ij}]} H_{ij}(l) - H_j \geq 0 \text{ 且 } \max_{l \in [0,L_{ij}]} H_{ij}(l) - H_i \geq 0 \\ [0,t_1], & \max_{l \in [0,L_{ij}]} H_{ij}(l) - H_j \geq 0 \text{ 且 } \max_{l \in [0,L_{ij}]} H_{ij}(l) - H_i \leq 0 \\ [1,1], & \max_{l \in [0,L_{ij}]} H_{ij}(l) - H_j \leq 0 \text{ 且 } \max_{l \in [0,L_{ij}]} H_{ij}(l) - H_i \geq 0 \\ [t_1,t_2], & other \end{cases}$$

$t_1, t_2 \in (0,1)$ is a quantitative parameter, $t_1 < t_2$ is expressed by the number of intervals, $L_{ij}$ is the linear distance between the blue tank target $T_j$ and the red tank $W_i$, and l is the horizontal distance from the actual point to the target $T_j$. $H_{ij\,(l)}$ is the distance between red and blue targets.$T_j$ represents the actual terrain height of l, $H_i$ and $H_j$ represent the actual terrain height of the positions of tank $W_i$ and tank $T_j$.

If the evaluation node and target are both fully visible and are above the mid-elevation, the terrain visibility threat degree is [$t_2$,1]; when neither can communicate, the threat degree is [0,0];if the red-sided $W_i$ is situated at a high point and higher than the middle elevation, and the blue-sided $T_j$ is situated at a low point and lower than the middle elevation, then the threat degree is [0,$t_1$];whenever the blue side of $T_j$ is higher than the middle elevation, and the red side of $W_i$ is lower than the middle elevation, the threat degree is [1,1]; As the shooting can be completed without a cross-view during indirect shooting, the degree of threat is uniformly set to [$t_1$,$t_2$]. By combining the target's position and elevation with the wargame environment, it is possible to evaluate the visibility of the target to the red tank $W_i$ by combining the visibility of target with environment.

### 3.5 Quantification of environmental indicators

The effects of the wargame confrontation environment are an important determinant of combat effectiveness. The better the environmental indicators, the better the equipment effectiveness of the chess pieces, and the worse the environmental indicators, the worse the combat effectiveness of the chess pieces[22-23]. Due to the fact that the simulation is carried out in a wargame environment, this article extracts the two major influencing factors for the wargame environment, urban residential areas and highways, for the quantification of environmental indicators.Tanks and infantry personnel can be concealed in urban residential areas so that the enemy cannot detect our targets, which is conducive to our defense. Highways can speed up the chess pieces' moving speed, and there are first-level and second-level highways in the wargame environment. The higher the level of the road, the better the chess pieces' speed.Therefore, environmental indicators can be quantified according to the environment where chess pieces are located, considering surrounding urban residential areas and road conditions. Determine whether there are first-class roads, second-class roads and residential areas in the two squares around the red tank $W_i$. Gaining threat degree $Te_i$ through judgment.

$$Te_i = w_1 h_1 + w_2 h_2 + w_3 r$$

$h_1$, $h_2$, and r represent first-class roads, second-class roads, and urban residential areas, respectively, and $w_1$, $w_2$, and $w_3$ represent weight vectors. If the above terrain environment is located around the chess piece, the associated value is assigned.

### 3.6 Quantification of target defense

Generally, the defense quantification of chess pieces includes the armor protection attributes. The different types of armor include composite armor, heavy armor, medium armor, light armor, and unarmored. This article assigns different armors different defense values, based on their armor protection capabilities.The table 2 below shows.

Table 2 Quantified value of target defense

| Armored attributes | Quantified value |
|---|---|
| composite armor | 1 |
| heavy armor | 0.7 |
| medium armor | 0.5 |
| light armor | 0.3 |
| unarmored | 0 |

4、Establishment of a multi-attribute quantitative threat model based on intuitionistic fuzzy numbers

By using the interval number method, this article indicates whether visibility is possible, and different threats are generated. Nevertheless, the quantified values of other threat targets are real numbers. To unify the problem solving method, this paper converts all interval numbers and real numbers to intuitionistic fuzzy numbers, and calculates the size of the threat by calculating the intuitionistic fuzzy numbers.

**4.1 Principles of conversion between different description forms**

The quantitative multi-attribute index set represented by different forms is X, and the quantitative index is x, $x \in X$ ; the intuitionistic fuzzy set is Y, and the intuitionistic fuzzy number of the index is y, $y \in Y$ . When converting indicators of different representations, considering the following principles should be followed in order to ensure that the form of the conversion is scientific and reasonable:

1) Range limitations

In mapping $x \to y$ , if $x \in X$ is present, then $y \in Y$ .

2) Characteristics of boundary

If x is the upper bound of its representation, then $y = \langle 1, 0 \rangle$ . And if x is the lower bound of its representation, then $y = \langle 0, 1 \rangle$ .

3) Monotonic map

When $x_1 < x_2$ , then $y_1 < y_2$ ; if $x_1 = x_2$ , then $y_1 = y_2$ , if $x_1 > x_2$ , then $y_1 > y_2$ .

**4.2 The interval number is converted into an intuitionistic fuzzy number**

During the wargame confrontation, it is difficult to draw accurate conclusions about the level of visibility. This article, therefore, uses the interval number method to express this type of difficult-to-determine information. To uniformly express that all the numbers in this article are transformed into intuitionistic fuzzy numbers for representation, here is a conversion method between interval numbers and intuitionistic fuzzy numbers. The interval number $\tilde{a}_i = \left[a_i^L, a_i^U\right]$ is transformed into intuitionistic fuzzy number $f_i = \langle \mu_i, \nu_i \rangle$ using the following formula.

$$\begin{cases} \mu_i = \dfrac{a_i L}{\max_{i=1,2,\ldots,n}\{a_i^U\}} \\ \nu_i = 1 - \dfrac{a_i^U}{\max_{i=1,2,\cdots,n}\{a_i^U\}} \end{cases}$$

Proof: 1) Range restrictions

If $\tilde{a}_i = \left[a_i^L, a_i^U\right] \in R$, then $\mu_i = \dfrac{a_i^L}{\max_{i=1,2,,n}\{a_i^U\}}$, $\mu_i = \dfrac{a_i^L}{\max_{i=1,2,\cdots,n}\{a_i^U\}} \in [0,1]$,

$\nu_i = 1 - \dfrac{a_i^U}{\max_{i=1,2,\cdots,n}\{a_i^U\}} \in [0,1]$   $0 \leq \mu_i + \nu_i = 1 - \dfrac{a_i^U - a_i^L}{\max_i\{a_i^U\}} \leq 1$

so

2) Boundary characteristics

$\tilde{a}_i$ takes the lower bound $[0,0]$, and $\max_{i=1,2,\cdots,n}\{a_i^U\} \neq 0$. Apparently

$\mu_i = \dfrac{a_i^L}{\max_{i=1,2,\cdots,n}\{a_i^U\}} = 0$ follows.

3) Mapping monotonic

Set $\tilde{a}_1 = \left[a_1^L, a_1^U\right]$ and $\tilde{a}_2 = \left[a_2^L, a_2^U\right]$ intervals. The degree of uncertainty between these numbers is usually expressed as a probability relation. In order to judge the monotonicity of the mapping, it is important to use the determined relationship (possibility is 1).So, when determining the size relationship, this article is mainly concerned with three situations: $\tilde{a}_1$ is smaller than $\tilde{a}_2$ ($\tilde{a}_1 < \tilde{a}_2$) by the full amount, $\tilde{a}_1$ equal $\tilde{a}_2$ ($\tilde{a}_1 = \tilde{a}_2$) and $\tilde{a}_1$ is completely greater than $\tilde{a}_2$ ($\tilde{a}_1 > \tilde{a}_2$)

A）If $\tilde{a}_1 < \tilde{a}_2$ and $a_1^L \leq a_1^U < a_2^L \leq a_2^U$, then

$$\mu_1 - \mu_2 = \frac{a_1 L}{\max_{i=1,2,\cdots,n}\{a_i^U\}} - \frac{a_2 L}{\max_{i=1,2,\cdots,n}\{a_i^U\}} < 0$$

$$v_1 - v_2 = 1 - \frac{a_1^U}{\max_{i=1,2,\cdots,n}\{a_i^U\}} - \left(1 - \frac{a_2^U}{\max_{i=1,2,\cdots,n}\{a_i^U\}}\right) = \frac{a_2^U - a_1^U}{\max_{i=1,2,\cdots,n}\{a_i^U\}} > 0$$

So $f_1 < f_2$

B) If $\tilde{a}_1 = \tilde{a}_2$, obviously $\mu_1 = \mu_2$, $v_1 = v_2$, so $f_1 = f_2$

C) If $\tilde{a}_1 > \tilde{a}_2$, and $a_2^L \leq a_2^U < a_1^L \leq a_1^U$. Then

$$\mu_1 - \mu_2 = \frac{a_1 L}{\max_{i=1,2,\cdots,n}\{a_i^U\}} - \frac{a_2^L}{\max_{i=1,2,\cdots,n}\{a_i^U\}} > 0$$

$$v_1 - v_2 = 1 - \frac{a_1 U}{\max_{i=1,2,\cdots,n}\{a_i^U\}} - \left(1 - \frac{a_2^U}{\max_{i=1,2,\cdots,n}\{a_i^U\}}\right) = \frac{a_2^U - a_1^U}{\max_{i=1,2,\cdots,n}\{a_i^U\}} < 0$$

So $f_1 > f_2$。

### 4.3 Convert real numbers into intuitionistic fuzzy numbers

To unify the index representation form, the real number type index representation must also be converted into intuitionistic fuzzy numbers. The following is the formula for calculating the membership degree and non-membership degree of the conversion of real numbers into IFN.

Beneficial index

$$\begin{cases} \mu_i = \beta \dfrac{a_i}{\max_{i=1,2,\cdots,n}\{a_i\}} \\ v_i = \beta \left(1 - \dfrac{a_i}{i=1,2,\cdots,n} \max\{a_i\}\right) \end{cases}$$

The cost index

$$\begin{cases} \mu_i = \beta \dfrac{\min_{i=1,2,\cdots,n}\{a_i\}}{a_i} \\ v_i = \beta \left(1 - \dfrac{\min i=1,2,\cdots,n\{a_i\}}{a_i}\right) \end{cases}$$

In the formula, real numbers are considered special cases of interval numbers. The function of factor $\beta \in [0,1]$ is to prevent the deterministic characteristics of real numbers from affecting other uncertainty indicators. The proof: can treat real numbers as special cases of interval numbers, so it can prove that real numbers are converted into intuitionistic fuzzy numbers via the same method used to prove interval numbers are converted into direct fuzzy numbers.

**4.4 Quantitative calculation of fuzzy numbers intuitively based on multiple attributes**

（1）This intuitionistic fuzzy entropy describes the degree of fuzzy judgment information provided by an intuitionistic fuzzy set. The larger the intuitionistic fuzzy entropy of an evaluation criterion, the smaller the weight should be given; otherwise, the larger needs to be. Based on formulas from the literature[24], we calculated the entropy weights for each intuitionistic fuzzy. Among them, ideal solution $S_i^+$ is a conceived optimal solution (scheme), and its attribute values hit the best value among the alternatives; and the negative ideal solution $S_i^-$ is the worst conceived solution (scheme), and its attribute values hit the worst value among the alternatives. $p_i$ is generated by comparing each alternative scheme with the ideal solution and negative ideal solution. If one of the solutions is closest to the ideal solution, but at the same time far from the negative ideal solution, then it is the best solution among the alternatives.

$$H_j = -\frac{1}{n\ln 2}\sum_{i=1}^{m}\left[\mu_{ij}\ln\mu_{ij} + v_{ij}\ln v_{ij} - (\mu_{ij}+v_{ij})\ln(\mu_{ij}+v_{ij}) - (1-\mu_{ij}-v_{ij})\ln 2\right]$$

If $\mu_{ij}=0$, $v_{ij}=0$, $\mu_{ij}+v_{ij}=0$, then $\mu_{ij}\ln\mu_{ij}=0$

$v_{ij}\ln v_{ij}=0$, $(\mu_{ij}+v_{ij})\ln(\mu_{ij}+v_{ij})=0$

The entropy weight of the j attribute is defined as:

$$w_j = \frac{1-H_j}{n-\sum_{j=1}^{n}H_j}$$

Among $w_j \geq 0, j=1,2,\cdots,n, \sum_{j=1}^{n}w_j = 1$

（2）Determine the optimal solution A+ and the worst solution A- using the following formula:

$$\begin{cases} A^+ = \{\langle \mu_1^+, v_1^+ \rangle, \langle \mu_2^+, v_2^+ \rangle, \cdots, \langle \mu_n^+, v_n^+ \rangle\} \\ A^- = \{\langle \mu_1^-, v_1^- \rangle, \langle \mu_2^-, v_2^- \rangle, \cdots, \langle \mu_n^-, v_n^- \rangle\} \end{cases}$$

In which

$$\mu_i^+ = \max_{j=1,2,\ldots,m} \{\mu_{ij}\}, v_i^+ = \min_{j=1,2,\ldots,m} \{v_{ij}\}$$

$$\mu_i^- = \min_{j=1,2,\cdots,m} \{\mu_{ij}\}, v_i^- = \max_{j=1,2,\cdots,m} \{v_{ij}\}$$

（3）Calculate the similarity between the fuzzy intuitionistic A and B as follows:

$$s(\langle \mu_1, v_1 \rangle, \langle \mu_2, v_2 \rangle) = 1 - \frac{|2(\mu_1 - \mu_2) - (v_1 - v_2)|}{3} \times \left(1 - \frac{\pi_1 + \pi_2}{2}\right) \frac{|2(v_1 - v_2) - (\mu_1 - \mu_2)|}{3} \times \left(\frac{\pi_1 + \pi_2}{2}\right)$$

In which, $\pi_1 = 1 - \mu_1 - v_1, \pi_2 = 1 - \mu_2 - v_2$

（4）Calculate the similarity $S_i^+$ and $S_i^-$ between each solution and the optimal solution and the worst solution based on the following formula:

$$\begin{cases} S_i^+ = \sum_{k=1}^n w_k \cdot s(\langle \mu_k^+, v_k^+ \rangle, \langle \mu_{ik}, v_{ik} \rangle) \\ S_i^- = \sum_{k=1}^n w_k \cdot s(\langle \mu_k^-, v_k^- \rangle, \langle \mu_{ik}, v_{ik} \rangle) \end{cases}$$

(5) Then calculate the relative closeness

$$p_i = S_i^- / (S_i^+ + S_i^-)$$

Comparing threat levels of opponents based on their closeness to the target depends on the level of threat assessment performed.

## 5、Multi-attribute threat quantitative simulation experiment using intuitionistic fuzzy numbers

The threat assessment problem is transformed into a multi-attribute decision-making problem, while the combat intention of the target is incorporated into the evaluation system to make the evaluation system more reasonable and the results more reliable. A simulation scene depicts ten tanks of the red and blue sides fighting each other, and ten opposite are found as chess pieces in the wargame. By using the war game deduction environment, each tank piece's attribute data will be derived in real time, including the opposite piece's position (hexagon number), elevation, type, distance, range, strike power and control point distance, as well as armor thickness. A table 3 is shown below.

Table 3 Data about each piece in wargaming

| Time | Number of target | Posting | Elevation | Type | Distance | Range | Taking Fire | Control point's distance | Thickness of armor |
|---|---|---|---|---|---|---|---|---|---|
| t1 | 1 | (15, 40) | 110 | Tank | 34 | 16 | 5 | 18 | Unarmored |
|  | 2 | (16, 41) | 110 | Tank | 34 | 20 | 2 | 19 | Unarmored |
|  | 3 | (17, 41) | 110 | Tank | 34 | 17 | 3 | 20 | Light armor |
|  | 4 | (18, 40) | 110 | Tank | 33 | 17 | 5 | 19 | Medium armor |
|  | 5 | (18, 41) | 110 | Tank | 34 | 20 | 3 | 20 | Composite armor |
|  | 6 | (17, 40) | 110 | Tank | 33 | 20 | 4 | 19 | Unarmored |
|  | 7 | (16, 40) | 110 | Tank | 33 | 20 | 5 | 18 | Unarmored |
|  | 8 | (15, 41) | 110 | Tank | 35 | 16 | 4 | 19 | Light armor |
|  | 9 | (18, 42) | 110 | Tank | 35 | 16 | 2 | 21 | Composite armor |
|  | 10 | (17, 42) | 110 | Tank | 35 | 17 | 3 | 21 | Heavy armor |
| t2 | 1 | (15, 39) | 110 | Tank | 32 | 16 | 5 | 17 | Unarmored |
|  | 2 | (16, 40) | 110 | Tank | 32 | 20 | 2 | 18 | Unarmored |
|  | 3 | (17, 41) | 110 | Tank | 33 | 17 | 3 | 20 | Light armor |
|  | 4 | (18, 40) | 110 | Tank | 32 | 17 | 5 | 19 | Medium armor |
|  | 5 | (18, 40) | 110 | Tank | 32 | 20 | 3 | 19 | Composite armor |
|  | 6 | (17, 40) | 110 | Tank | 32 | 20 | 4 | 19 | Unarmored |
|  | 7 | (16, 39) | 110 | Tank | 31 | 20 | 5 | 17 | Unarmored |
|  | 8 | (15, 41) | 110 | Tank | 34 | 16 | 4 | 19 | Light armor |
|  | 9 | (18, 41) | 110 | Tank | 33 | 16 | 2 | 20 | Composite armor |
|  | 10 | (17, 41) | 110 | Tank | 33 | 17 | 3 | 20 | Heavy armor |
| t3 | 1 | (15, 39) | 110 | Tank | 31 | 16 | 5 | 17 | Unarmored |
|  | 2 | (16, 40) | 110 | Tank | 31 | 20 | 2 | 18 | Unarmored |
|  | 3 | (17, 41) | 110 | Tank | 32 | 17 | 3 | 20 | Light armor |
|  | 4 | (18, 40) | 110 | Tank | 31 | 17 | 5 | 19 | Medium armor |
|  | 5 | (18, 40) | 110 | Tank | 31 | 20 | 3 | 19 | Composite armor |
|  | 6 | (17, 40) | 110 | Tank | 31 | 20 | 4 | 19 | Unarmored |
|  | 7 | (16, 39) | 110 | Tank | 30 | 20 | 5 | 17 | Unarmored |
|  | 8 | (15, 41) | 110 | Tank | 33 | 16 | 4 | 19 | Light armor |
|  | 9 | (18, 40) | 110 | Tank | 31 | 16 | 2 | 19 | Composite armor |
|  | 10 | (17, 41) | 110 | Tank | 32 | 17 | 3 | 20 | Heavy armor |
| t4 | 1 | (15, 39) | 110 | Tank | 31 | 16 | 5 | 17 | Unarmored |
|  | 2 | (16, 39) | 110 | Tank | 30 | 20 | 2 | 17 | Unarmored |
|  | 3 | (17, 41) | 110 | Tank | 33 | 17 | 3 | 20 | Light armor |
|  | 4 | (18, 39) | 110 | Tank | 31 | 17 | 5 | 18 | Medium armor |
|  | 5 | (18, 40) | 110 | Tank | 32 | 20 | 3 | 19 | Composite armor |
|  | 6 | (17, 39) | 110 | Tank | 31 | 20 | 4 | 18 | Unarmored |

| | 7 | (16, 39) | 110 | Tank | 30 | 20 | 5 | 17 | Unarmored |
| --- | --- | --- | --- | --- | --- | --- | --- | --- | --- |
| | 8 | (15, 41) | 110 | Tank | 33 | 16 | 4 | 19 | Light armor |
| | 9 | (18, 40) | 110 | Tank | 32 | 16 | 2 | 19 | Composite armor |
| | 10 | (17, 41) | 110 | Tank | 33 | 17 | 3 | 20 | Heavy armor |
| t5 | 1 | (15, 39) | 110 | Tank | 31 | 16 | 5 | 17 | Unarmored |
| | 2 | (16, 38) | 110 | Tank | 30 | 20 | 2 | 16 | Unarmored |
| | 3 | (17, 40) | 110 | Tank | 33 | 17 | 3 | 19 | Light armor |
| | 4 | (18, 38) | 110 | Tank | 31 | 17 | 5 | 17 | Medium armor |
| | 5 | (18, 40) | 110 | Tank | 33 | 20 | 3 | 19 | Composite armor |
| | 6 | (17, 39) | 110 | Tank | 32 | 20 | 4 | 18 | Unarmored |
| | 7 | (16, 39) | 110 | Tank | 31 | 20 | 5 | 17 | Unarmored |
| | 8 | (15, 41) | 110 | Tank | 33 | 16 | 4 | 19 | Light armor |
| | 9 | (18, 40) | 110 | Tank | 33 | 16 | 2 | 19 | Composite armor |
| | 10 | (17, 41) | 110 | Tank | 34 | 17 | 3 | 20 | Heavy armor |

By integrating real numbers and interval numbers to form intuitionistic fuzzy number formulas, the target distance is quantified according to the distance threat quantification method in Section 3.1, and the distance threat degree is then turned into a fuzzy number using the conversion method of real numbers and fuzzy numbers in Section 4. The calculated result for the target distance threat degree is provided in the table 4. Consider the threat measurement at T1 as an example of a decision matrix.

Table 4  Calculation of target distance threat

| Target | Value of normal terrain stamina consumption | Value of special terrain energy consumption | Distance between the blue tank and the control point | Consider the maximum distance between two borders | Distance between tanks i and j | Representation of real numbers | Representation of fuzzy numbers intuitively |
| --- | --- | --- | --- | --- | --- | --- | --- |
| 1 | 3 | 6 | 18 | 50 | 34 | 8.16 | [0.187378998, 0.012621002] |
| 2 | 3 | 6 | 19 | 50 | 34 | 8.155 | [0.18749387, 0.01250613] |
| 3 | 3 | 6 | 20 | 50 | 34 | 8.15 | [0.187608882, 0.012391118] |
| 4 | 3 | 6 | 19 | 50 | 33 | 8.655 | [0.176663586, 0.023336414] |
| 5 | 3 | 6 | 20 | 50 | 34 | 8.15 | [0.187608882, 0.012391118] |
| 6 | 3 | 6 | 19 | 50 | 33 | 8.655 | [0.176663586, 0.023336414] |
| 7 | 3 | 6 | 18 | 50 | 33 | 8.66 | [0.176561598, 0.023438402] |

| 8 | 3 | 6 | 19 | 50 | 35 | 7.655 | [0.199738767, 0.000261233] |
| 9 | 3 | 6 | 21 | 50 | 35 | 7.645 | [0.2, 0.0] |
| 10 | 3 | 6 | 21 | 50 | 35 | 7.645 | [0.2, 0.0] |

Based on the real number conversion intuitionistic fuzzy number formula, the target speed is quantified according to the 3.2 section speed threat quantification method, and the speed threat degree is converted into an intuitionistic fuzzy number according to the 4th section of the conversion method of real numbers and intuitionistic fuzzy numbers. Calculation of target velocity threat degree. Create a matrix showing the target velocity threat measurement at T1, for example, as shown in the table 5.

Table 5 Calculation of target speed threat

| Target | The target speed | The relative speed between the opposite and the us | The real number representation | Intuitive fuzzy number representation |
|---|---|---|---|---|
| 1 | 125 | 325 | 2.6 | [0.153863899, 0.046136101] |
| 2 | 150 | 350 | 2.333333333 | [0.171440811, 0.028559189] |
| 3 | 200 | 400 | 2 | [0.2, 0.0] |
| 4 | 200 | 400 | 2 | [0.2, 0.0] |
| 5 | 200 | 400 | 2 | [0.2, 0.0] |
| 6 | 200 | 400 | 2 | [0.2, 0.0] |
| 7 | 200 | 400 | 2 | [0.2, 0.0] |
| 8 | 150 | 350 | 2.333333333 | [0.171440811, 0.028559189] |
| 9 | 175 | 375 | 2.142857143 | [0.186672886, 0.013327114] |
| 10 | 150 | 350 | 2.333333333 | [0.171440811, 0.028559189] |

The real number conversion intuitionistic fuzzy number formula is applied to determine the target's attack ability using the attack threat quantification method in Section 3.3, and then the attack threat degree is converted into an intuitionistic fuzzy number using the conversion method of real numbers and intuitionistic fuzzy numbers in Section 4. The calculated solution result of the target attack threat. Determine the decision matrix using the target attack threat measurement at T1 as an example, as shown in the table 6.

Table 6 Calculation of target attack threat

| Target | Manoeuvrability | Weapon system attack capability | Reconnaissance capability | Capability of indirect shots | The ammunition carrying capacity | ECM capability | Offensive missile capability | The real number representation | Intuitive fuzzy number representation |
|---|---|---|---|---|---|---|---|---|---|
| 1 | 6 | {'type1': 1, 'type2': 0.5, 'type3': 1.5} | {'type1': 1, 'type2': 0.5, 'type3': 1.5} | 1 | 3 | 1 | 1 | 10.8730228 | [0.2, 0.0] |
| 2 | 6 | {'type1': 1, 'type2': 0.5, 'type3': 1.5} | {'type1': 1, 'type2': 0.5, 'type3': 1.5} | 1 | 3 | 1 | 1 | 10.8730228 | [0.2, 0.0] |
| 3 | 6 | {'type1': 1, 'type2': 0.5, 'type3': 1.5} | {'type1': 1, 'type2': 0.5, 'type3': 1.5} | 1 | 3 | 1 | 1 | 10.8730228 | [0.2, 0.0] |
| 4 | 6 | {'type1': 1, 'type2': 0.5, 'type3': 1.5} | {'type1': 1, 'type2': 0.5, 'type3': 1.5} | 1 | 3 | 1 | 1 | 10.8730228 | [0.2, 0.0] |
| 5 | 6 | {'type1': 1, 'type2': 0.5, 'type3': 1.5} | {'type1': 1, 'type2': 0.5, 'type3': 1.5} | 1 | 3 | 1 | 1 | 10.8730228 | [0.2, 0.0] |
| 6 | 6 | {'type1': 1, 'type2': 0.5, 'type3': 1.5} | {'type1': 1, 'type2': 0.5, 'type3': 1.5} | 1 | 3 | 1 | 1 | 10.8730228 | [0.2, 0.0] |
| 7 | 6 | {'type1': 1, 'type2': 0.5, 'type3': 1.5} | {'type1': 1, 'type2': 0.5, 'type3': 1.5} | 1 | 3 | 1 | 1 | 10.8730228 | [0.2, 0.0] |
| 8 | 6 | {'type1': 1, 'type2': 0.5, 'type3': 1.5} | {'type1': 1, 'type2': 0.5, 'type3': 1.5} | 1 | 3 | 1 | 1 | 10.8730228 | [0.2, 0.0] |
| 9 | 6 | {'type1': 1, 'type2': 0.5, 'type3': 1.5} | {'type1': 1, 'type2': 0.5, 'type3': 1.5} | 1 | 3 | 1 | 1 | 10.8730228 | [0.2, 0.0] |
| 10 | 6 | {'type1': 1, 'type2': 0.5, 'type3': 1.5} | {'type1': 1, 'type2': 0.5, 'type3': 1.5} | 1 | 3 | 1 | 1 | 10.8730228 | [0.2, 0.0] |

By combining the interval number conversion intuitionistic fuzzy number formula with the terrain visibility threat quantification method in section 3.4, the target visibility is quantified, and then the terrain visibility threat degree is converted into intuitionistic fuzzy numbers. Solution result of the visibility threat degree of the target terrain. Create a decision matrix using the metric of the visibility threat of T1 as an example, as shown in the table 7.

Table 7 Calculated threat level for terrain visibility

| Target | The red square elevation | The blue square elevation | The highest altitude between red and blue | The interval number representation | Intuitive fuzzy number representation |
|---|---|---|---|---|---|
| 1 | 130 | 110 | 150 | [0, 0.2] | [0.0, 0.0] |
| 2 | 130 | 110 | 150 | [0, 0.2] | [0.0, 0.0] |
| 3 | 130 | 110 | 150 | [0, 0.2] | [0.0, 0.0] |
| 4 | 130 | 110 | 150 | [0, 0.2] | [0.0, 0.0] |
| 5 | 130 | 110 | 150 | [0, 0.2] | [0.0, 0.0] |
| 6 | 130 | 110 | 150 | [0, 0.2] | [0.0, 0.0] |
| 7 | 130 | 110 | 150 | [0, 0.2] | [0.0, 0.0] |
| 8 | 130 | 110 | 150 | [0, 0.2] | [0.0, 0.0] |
| 9 | 130 | 110 | 150 | [0, 0.2] | [0.0, 0.0] |
| 10 | 130 | 110 | 150 | [0, 0.2] | [0.0, 0.0] |

As described in section 3.5, environmental indicators are quantified and converted into intuitionistic fuzzy numbers using the real number conversion intuitionistic fuzzy number formula, then converted into real numbers using the conversion method of real numbers and intuitionistic fuzzy numbers in section 4. The calculation result of the target environment index. Construct a decision matrix using the environmental indicator threat measurement at T1 as an example, as shown in the table 8.

Table 8 Calculation of the target environment index

| Target | The first class road (weight $W_1$) | The secondary road (weight $W_2$) | The urban residential area (weight $W_3$) | The real number representation | Intuitive fuzzy number representation |
|---|---|---|---|---|---|
| 1 | 0(3) | 0(2) | 0(1) | 0 | [0.2, 0.0] |
| 2 | 0(3) | 0(2) | 0(1) | 0 | [0.2, 0.0] |
| 3 | 0(3) | 0(2) | 0(1) | 0 | [0.2, 0.0] |
| 4 | 0(3) | 0(2) | 0(1) | 0 | [0.2, 0.0] |
| 5 | 0(3) | 0(2) | 0(1) | 0 | [0.2, 0.0] |
| 6 | 0(3) | 0(2) | 0(1) | 0 | [0.2, 0.0] |
| 7 | 0(3) | 0(2) | 0(1) | 0 | [0.2, 0.0] |
| 8 | 0(3) | 0(2) | 0(1) | 0 | [0.2, 0.0] |
| 9 | 1(3) | 0(2) | 0(1) | 3 | [6.6644e-05, 0.199933356] |
| 10 | 0(3) | 0(2) | 0(1) | 0 | [0.2, 0.0] |

Using the real number conversion intuitionistic fuzzy number formula, the target defense is quantified following the target defense quantification method in Section 3.6, and then the target defense quantified value is converted into intuitionistic fuzzy numbers following the conversion method of real numbers and intuitionistic fuzzy numbers in Section 4. Calculate the quantitative results of target defense in the table. Build a decision matrix by using the quantification of defense indicators at T1 as an example, as shown in the table 9.

Table 9 Calculation of the target defense quantitatively

| Target | Armored type | The real number representation | Intuitive fuzzy number representation |
|---|---|---|---|
| 1 | Unarmored | 0 | [0.2, 0.0] |

| 2 | Unarmored | 0 | [0.2, 0.0] |
|---|---|---|---|
| 3 | Light armor | 0.3 | [0.000664452, 0.199335548] |
| 4 | Medium armor | 0.5 | [0.000399202, 0.199600798] |
| 5 | Composite armor | 1 | [0.0001998, 0.1998002] |
| 6 | Unarmored | 0 | [0.2, 0.0] |
| 7 | Unarmored | 0 | [0.2, 0.0] |
| 8 | Light armor | 0.3 | [0.000664452, 0.199335548] |
| 9 | Composite armor | 1 | [0.0001998, 0.1998002] |
| 10 | Heavy armor | 0.7 | [0.000285307, 0.199714693] |

A unified intuitiveistic fuzzy number representation has been created for all multi-attribute indicators. An example of an intuitionistic fuzzy number representation of threat assessment indicators is illustrated in the following table 10.

Table 10 Information decision table for threat target parameters (intuitionistic fuzzy number)

| | Tank1 | Tank2 | Tank3 | Tank4 | Tank5 | Tank6 | Tank7 | Tank8 | Tank9 | Tank10 |
|---|---|---|---|---|---|---|---|---|---|---|
| Quantification of target distance threats | [0.187378998, 0.012621002] | [0.18749387, 0.01250613] | [0.187608882, 0.012391118] | [0.176663586, 0.023336414] | [0.187608882, 0.012391118] | [0.176663586, 0.023336414] | [0.176561598, 0.023438402] | [0.199738767, 0.000261233] | [0.2, 0.0] | [0.2, 0.0] |
| Quantification of target speed threats | [0.153863899, 0.046136101] | [0.171440811, 0.028559189] | [0.2, 0.0] | [0.2, 0.0] | [0.2, 0.0] | [0.2, 0.0] | [0.2, 0.0] | [0.171440811, 0.028559189] | [0.186672886, 0.013327114] | [0.171440811, 0.028559189] |
| Quantifying the threat from target attacks | [0.2, 0.0] | [0.2, 0.0] | [0.2, 0.0] | [0.2, 0.0] | [0.2, 0.0] | [0.2, 0.0] | [0.2, 0.0] | [0.2, 0.0] | [0.2, 0.0] | [0.2, 0.0] |
| Quantifying the threat posed by terrain visibility | [0.0, 0.0] | [0.0, 0.0] | [0.0, 0.0] | [0.0, 0.0] | [0.0, 0.0] | [0.0, 0.0] | [0.0, 0.0] | [0.0, 0.0] | [0.0, 0.0] | [0.0, 0.0] |
| Quantification of environmental indicators of threat | [0.2, 0.0] | [0.2, 0.0] | [0.2, 0.0] | [0.2, 0.0] | [0.2, 0.0] | [0.2, 0.0] | [0.2, 0.0] | [0.2, 0.0] | [6.6644e-05, 0.199933356] | [0.2, 0.0] |
| Quantification of target defense | [0.2, 0.0] | [0.2, 0.0] | [0.000664452, 0.199335548] | [0.000399202, 0.199600798] | [0.0001998, 0.1998002] | [0.2, 0.0] | [0.2, 0.0] | [0.000664452, 0.199335548] | [0.0001998, 0.1998002] | [0.000285307, 0.199714693] |

By obtaining data represented by the intuitionistic vagueness of the threat assessment indicators shown in the table, formulae in 4.4 may be used to obtain the intuitionistic vague target threat assessment based on multiple attribute decision-making approaches. As shown in the table 11, determine the target threat assessment results.

Table 11 Threat assessment for target

| | |
|---|---|
| $S_i^+$ | [0.9900131572106283, 0.9930194457658972, 0.9713249517102417, 0.9694274902547305, 0.9712630240082707, 0.9960298049584839, 0.9960124538670997, 0.9685356920167532, 0.9447732710194203, 0.9685296037271114] |
| $S_i^-$ | [0.9451975215527424, 0.9421912329974735, 0.963885727053129, 0.9657831885086402, 0.9639476547551001, 0.9391808738048868, 0.9391982248962711, 0.9666749867466174, 0.9904374077439504, 0.9666810750362593] |
| $P_i$ | [0.5115790069137391, 0.5131324752716081, 0.5019220710020746, 0.5009415775207532, 0.5018900705058751, 0.5146880470889931, 0.5146790810929003, 0.5004807500523212, 0.4882017660336315, 0.500477603991942] |
| Ranking | T6>T7>T2>T1>T3>T5>T4>T8>T10>T9 |

In the table 12, the opposite target at T1 is shown as a threat.

Table 12 Ranking of opposite targets at time T

| Type of piece | Indicator comprehensive | Ranking |
|---|---|---|
| Tank 1 | 0.511579007 | 4 |
| Tank 2 | 0.513132475 | 3 |
| Tank 3 | 0.501922071 | 5 |
| Tank 4 | 0.500941578 | 7 |
| Tank 5 | 0.501890071 | 6 |
| Tank 6 | 0.514688047 | 1 |
| Tank 7 | 0.514679081 | 2 |
| Tank 8 | 0.50048075 | 8 |
| Tank 9 | 0.488201766 | 10 |
| Tank 10 | 0.500477604 | 9 |

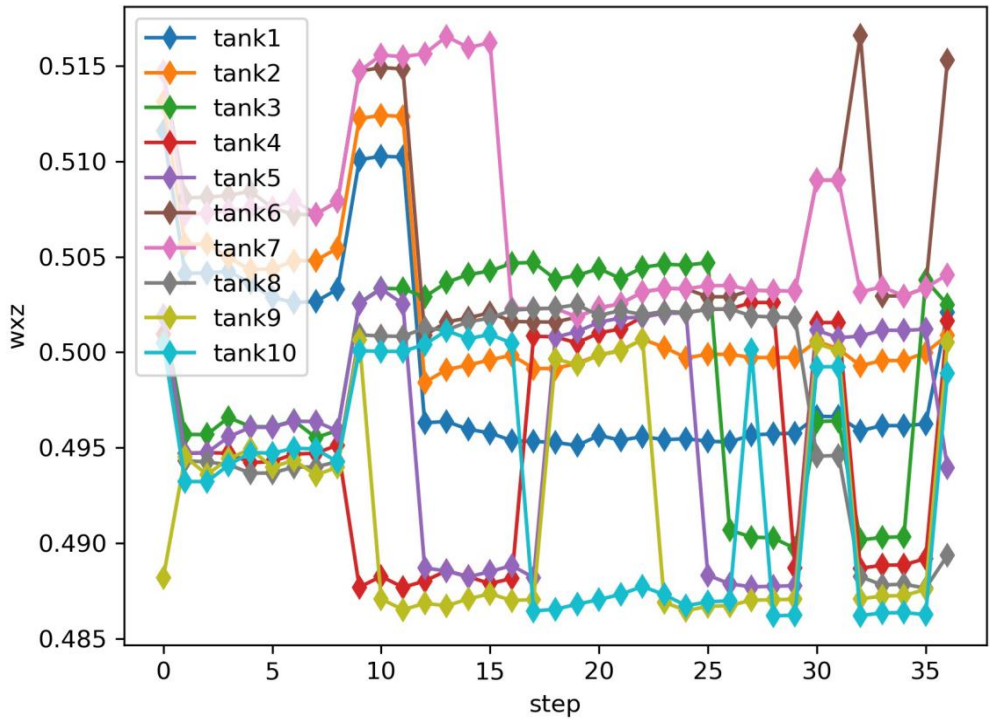

Figure 3    The threat value on the ordinate, and the threat of the opponent's ten tanks at time T represented by ten colors on the abscissa.

Based on the evaluation results, it can be concluded that the blue T6 tank is the most harmful and the T7 tank is the second most harmful, this is shown in figure 3. This article does not limit itself to subjective analysis of experts, but also introduces reinforcement learning, associates the reinforcement learning intelligent algorithm through the reward function and analyzes the scientific nature of the method through an improvement of the actual intelligent AI algorithm's winning rate.

## 6、Reinforcement learning and multi-attribute threat fusion model

### 6.1 Reinforcement learning algorithm and multi-attribute model formulation

Prior work in this article has described the quantied value of multi-attribute decision-making threat based on the entropy weight method. The subsequent chapters deal with how to associate this value with reinforcement learning. This part of the work is also the goal of this article. Its essence is to establish a multi-attribute decision-making mechanism that is based on reinforcement learning, and then select the entity with the highest threat to establish the return value and threat degree. The higher the threat degree, the greater the return value, this is shown in figure 4.

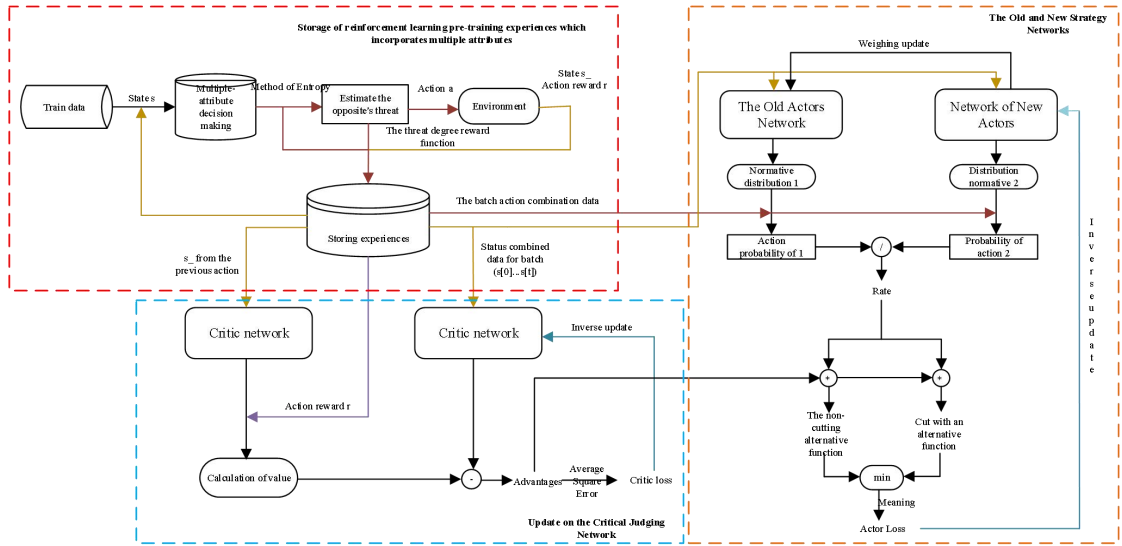

Figure 4  A fusion model of reinforcement learning and multi-attribute threat estimation based on AC framework.The module mainly consists of a reinforcement learning pre-training experience storage module that integrates multi-attribute decision-making, Critic evaluation network update module, and a new and old strategy network module

A reinforcement learning algorithm is built using the AC framework to achieve intelligent decision-making. It includes a reinforcement learning pre-training module that integrates multi-attribute decision-making, Critic evaluation network update module and a new and old strategy network update module.In the intensive learning pre-training experience storage module, multi-attribute decision making mainly uses state data obtained from the wargame environment, such as elevation, distance, armor thickness, etc., to make multi-attribute decisions. By normalizing the data, calculating the threat of each piece of the opponent by using the entropy method, and then setting the reward function and storing it in the experience store, further actions in the environment will be taken to obtain the next state and action rewards. The Critic network calculates the value from the reward value determined during the last step of the action. combines the experience store data with the value calculated by the Critic network, slashes it from the reward value determined during the last action, then returns to update the Critic network parameters.  As the advantage value guides the calculation of the actor network value, the network outputs the action value according to the old and new networks, and the distribution probability overall, and selects and outputs the action from the network. As a result, the advantage value is corrected, the actor loss is calculated, and the actor network is updated in the reverse direction.

## 6.2 Setting reward function value

As a core challenge of deep reinforcement learning in solving practical tasks, the sparse reward problem relates to the fact that the training environment cannot supervise the updating of agent parameters in the process of reinforcement learning [25]. When supervised learning is used, the training process is supervised by humans, while in reinforcement learning, rewards are used to supervise the training process, and the agent optimizes strategies based on rewards [19]. In the experimental environment in this article, the pawn will only transmit a victory message if it reaches the control point or annihilates the opposite pawn, or if it reaches the control point or if we are all annihilated. Both cases have no rewards at every step of the training process, that is, the sparse reward problem can affect algorithm convergence. So, this article tries to use multi-attribute decision-making to judge the eopposite's threat at every step, and then set the reward function according to the threat to get a higher reward value by hitting the threatening pieces.Furthermore, an additional reward mechanism is introduced. The reward value is determined according to the state of the chess piece and the distance from the control point. Additionally, in order to prevent the agent from falling into an optimal local situation, one reward is added every time the agent wins. A table of specific additional rewards is shown in Table 13.

Table 13 Additional award

| Situation | Reward |
| --- | --- |
| The state is now closer to the control point than the previous state | Reward+0.5 |
| This state is nearly as far from the control point as the previous state | Reward-0.3 |
| The map boundary has been reached | Reward-1 |
| Consumption per step (to avoid falling into local optimum) | Reward-0.005 |
| The opposite piece was hit | Reward+（5*Risk of being hit by a piece） |
| Hit by an opposite round | Reward-（5*Risk of being hit by a piece） |

| An opposite piece is annihilated | Reward+10 |
|---|---|
| Taking out one of the opposite's pieces will lead to victory | Reward+20 |
| Defeat an opposite piece leading to failure (other opposite pieces reach the control point) | Reward-10 |
| Get to the control point | Reward+10 |
| opposite wins | Reward-10 |

When the above additional rewards are added to the training process, the convergence speed can be significantly accelerated, and the likelihood that the agent falls into the local optimum is significantly reduced.

## 7、An experiment simulation of wargames

### 7.1 Experiment environment introduction

It is the objective of this article to conduct experimental verification in the self-developed wargame environment. Figure 5 shows the situation that exists during the initial red and blue confrontation deployment. There are two tank pawns on each side, and the center is the point of contention. In a confrontation, both sides compete for control points, and the party that reaches the middle red flag first wins. At the same time, both red and blue parties can shoot at each other, while they can hide in urban residential areas. By concealing, it is difficult for our opponents to find our targets. Each hexagon has its own number and elevation. The higher the elevation, the darker the hexagon. On the highway, the tanks move faster than on the secondary roads. The red straight line represents the secondary road and the black straight line represents the primary road. As the cross symbol represents aiming and shooting, the destroyed target disappears from the map.

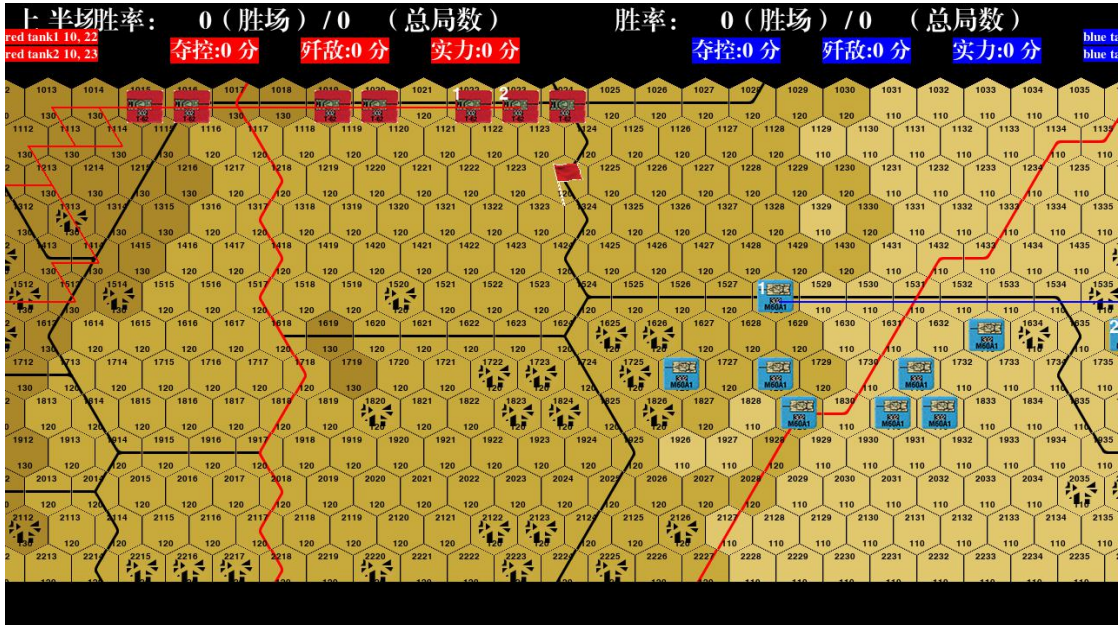

Figure 5　Gaming environment display. The red and blue pawns fight separately, the red flag in the middle is the control point, and the first player to reach the control point wins. When all the chess pieces on one side are destroyed, the other side wins.

## 7.2 Results and analysis of the experiment

　　In this article, the PPO algorithm and the PPO algorithm combined with multi-attribute decision-making are used to compare and analyze the winning rate. MADM-PPO and PPO are trained for 24 hours, and this article uses the MADM-PPO algorithm as the red side and the rule-based blue side algorithm to fight. At the same time, the second round uses the PPO algorithm as the red side, and the blue side fights according to rules. Next, this article observes the winning percentage of both algorithms in 100 games. Experiments have shown that the agents using the PPO reinforcement learning algorithm combined with the multi-attribute decision-making method performed better than the agents using the PPO algorithm based on the threat of the opponent. As can be seen in the figure 6 and figure 7, the threat-based multi-attribute decision-making method designed in this paper, combined with PPO algorithm of reinforcement learning, proves to effectively improve the effectiveness of intelligent wargame decision-making. A winning percentage chart is presented in the table 14, table 15.

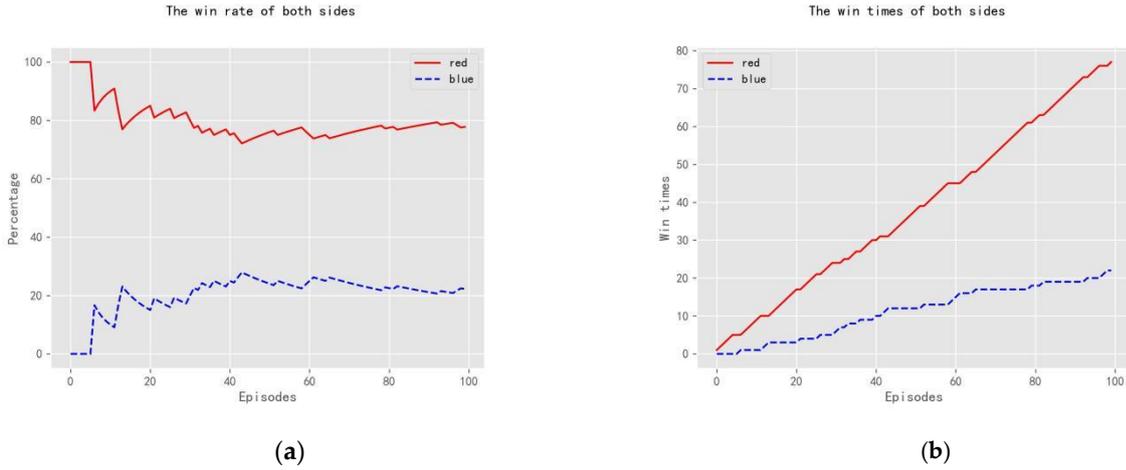

**Figure 6.** (**a**) Win rate: the red side is the AI of MADM-PPO intelligent algorithm and the blue side is rule-based AI; (**b**) Win times: the red side is the AI of MADM-PPO intelligent algorithm and the blue side is rule-based AI; The winning rate and the number of wins for the red and blue sides. The first round wins so one side starts from 1 and the other from 0.

**Table 14.** Comparison of the number of winning matches between red and blue teams after swapping positions.

| Algorithm | Victory Number | Rounds |
|---|---|---|
| MADM-PPO | 78 | 100 |
| Rule | 22 | 100 |

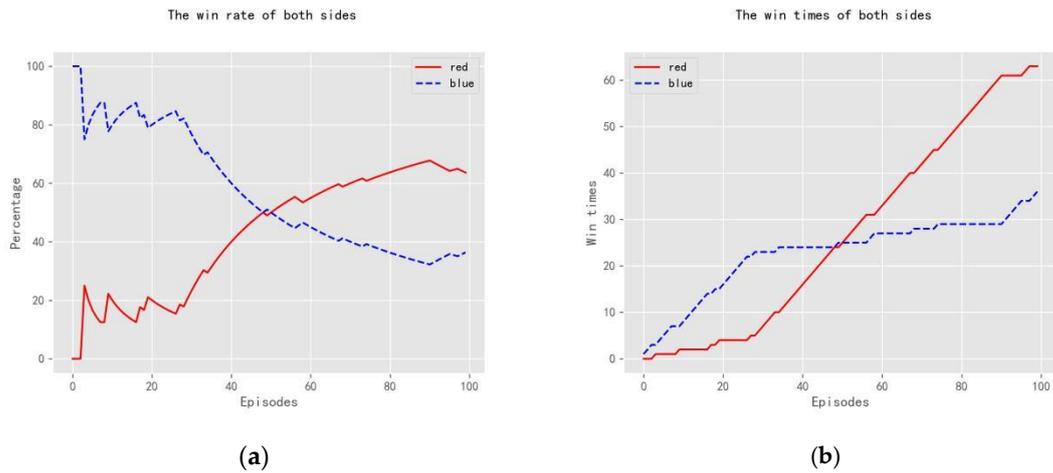

**Figure 7.** (**a**) Win rate: the red side is the AI of PPO intelligent algorithm and the blue side is rule-based AI; (**b**) Win times: the red side is the AI of PPO intelligent algorithm and the blue side is rule-based AI; The winning rate and the number of wins for the red and blue sides. The first round wins so one side starts from 1 and the other from 0.

Table 15. Comparison of winning matches between red and blue.

| Algorithm | Victory Number | Rounds |
|---|---|---|
| PPO | 62 | 100 |
| Rule | 38 | 100 |

The experimental results show that the MADM-PPO model can reduce the number of times to explore during training, and improve the problem that the PPO algorithm takes too long to train. It shows that the introduction of prior knowledge improves the performance of the PPO algorithm, and has a certain theoretical significance for improving the efficiency of the algorithm, the detail score is shown in Figure 8.

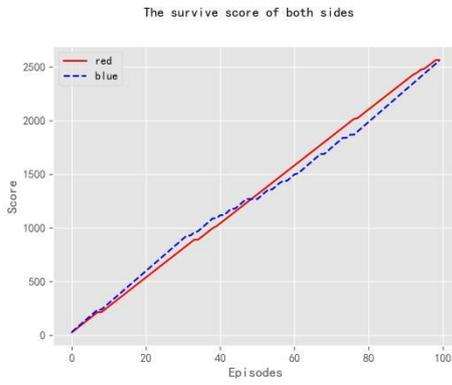

(a)

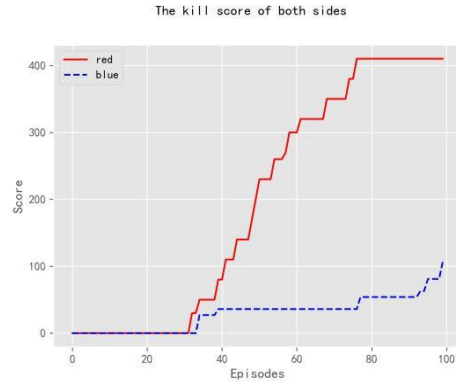

(b)

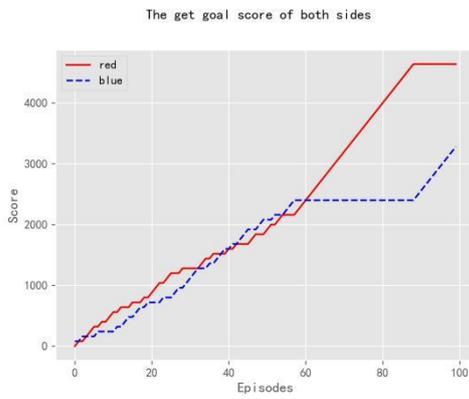

(c)

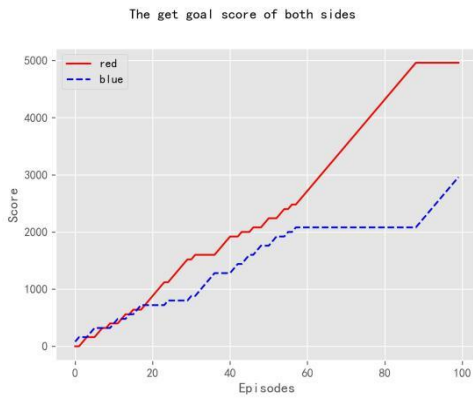

(d)

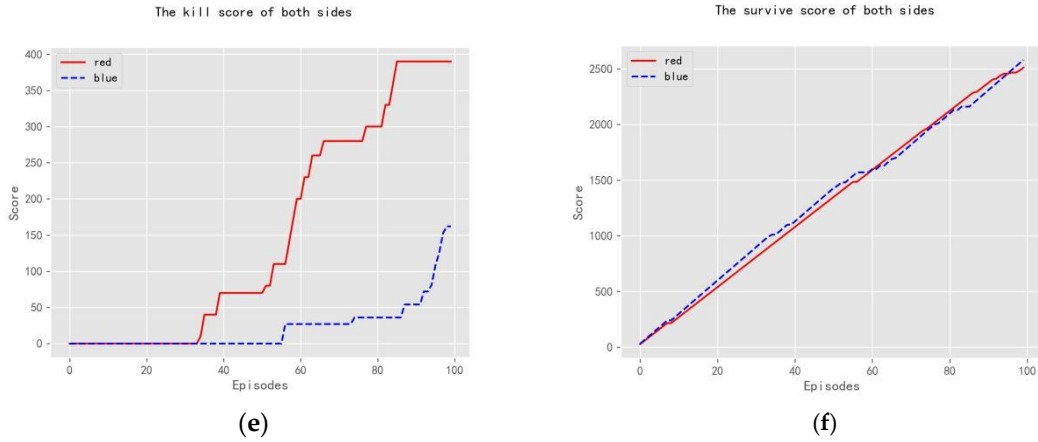

(e)                          (f)

**Figure 8.** (**a**) The get goal score of both sides (Red: PPO); (**b**) the kill score of both sides (Red: PPO); (**c**) the survive score of both sides(Red: PPO); (**d**) the get goal score of both sides (Red: MADM-PPO); (**e**) the kill score of both sides (Red: MADM-PPO); (**f**) the survive score of both sides (Red: MADM-PPO). The *x*-axis is the training episodes, and the *y*-axis is the score. Red and blue represent two teams in the confrontation environment.

## 8、Conclusion

Using the wargaming environment, the paper proposes an entity that combines multi-attribute decision-making and reinforcement learning to solve the problem that the reinforcement learning algorithm cannot quickly converge in war game training and the agent has a low winning rate against the algorithm.As part of this study, this paper conducts experiments on the multi-attribute decision-making and reinforcement learning algorithms in a wargame simulation environment, and obtains red and blue confrontation data from the wargame environment. Calculate the weight of each attribute based on the intuitionistic fuzzy number weight calculations. Then determine the threat posed by each opponent's chess pieces. On the basis of the degree of threat, the red side reinforcement learning reward function is constructed and the AC framework is trained with the reward function, and the algorithm combines multi-attribute decision-making with reinforcement learning. A study demonstrated that the algorithm can gradually increase the reward value of the agent when exploring an environment over a short training period, while the final victory rate of the agent against specific rules and strategies reached 78%, which is significantly higher than that of a pure reinforcement learning algorithm, which is 62%. Solved the convergence difficulties of the state-space wargame's sparse rewards caused by the randomization of an agent's neural network. For the algorithm design of intelligent wargaming, this is the first attempt in this field to combine the multi-attribute decision-making method in management with the reinforcement learning algorithm in cybernetics. An interdisciplinary approach to cross-innovation in academia could lead to improvements in the design of intelligent wargames and even improvements in reinforcement learning algorithms.

## Reference

[1] Pang Z J, Liu R Z, Meng Z Y, et al. On reinforcement learning for full-length game of starcraft[C]//Proceedings of the AAAI Conference on Artificial Intelligence. 2019, 33(01): 4691-4698.

[2] Silver D, Huang A, Maddison C J, et al. Mastering the game of Go with deep neural networks and tree search[J]. nature, 2016, 529(7587): 484-489.

[3] Ye D, Liu Z, Sun M, et al. Mastering complex control in moba games with deep reinforcement learning[C]//Proceedings of the AAAI Conference on Artificial Intelligence. 2020, 34(04): 6672-6679.

[4] Silver D, Schrittwieser J, Simonyan K, et al. Mastering the game of go without human knowledge[J]. nature, 2017, 550(7676): 354-359.

[5] Barriga N A, Stanescu M, Besoain F, et al. Improving rts game ai by supervised policy learning, tactical search, and deep reinforcement learning[J]. IEEE Computational Intelligence Magazine, 2019, 14(3): 8-18.

[6] Schrittwieser J, Antonoglou I, Hubert T, et al. Mastering atari, go, chess and shogi by planning with a learned model[J]. Nature, 2020, 588(7839): 604-609.

[7] Barriga N, Stanescu M, Buro M. Combining strategic learning with tactical search in real-time strategy games[C]//Proceedings of the AAAI Conference on Artificial Intelligence and Interactive Digital Entertainment. 2017, 13(1).

[8] Kong depeng, Chang Tianqing, Hao Na, Zhang Lei, Guo Libin. Multi attribute index processing method of ground combat target threat assessment [J]. Acta automatica Sinica, 2021,47 (01): 161-172

[9] Zhong S, Tan J, Dong H, et al. Modeling-Learning-Based Actor-Critic Algorithm with Gaussian Process Approximator[J]. Journal of Grid Computing, 2020, 18(2): 181-195.

[10] Littman M L. Reinforcement learning improves behaviour from evaluative feedback[J]. Nature, 2015, 521(7553): 445-451.

[11] Littman M L. Markov games as a framework for multi-agent reinforcement learning[M]//Machine learning proceedings 1994. Morgan Kaufmann, 1994: 157-163.

[12] Sutton R S, Barto A G. Reinforcement learning: An introduction[M]. MIT press, 2018.

[13] Hussain A, Chun J, Khan M. A novel multicriteria decision making (MCDM) approach for precise decision making under a fuzzy environment[J]. Soft Computing, 2021, 25(7): 5645-5661.

[14] Opricovic S, Tzeng G H. Compromise solution by MCDM methods: A comparative analysis of VIKOR and TOPSIS[J]. European journal of operational research, 2004, 156(2): 445-455.

[15] Kou G, Lu Y, Peng Y, et al. Evaluation of classification algorithms using MCDM and rank correlation[J]. International Journal of Information Technology & Decision Making, 2012, 11(01): 197-225.

[16] Von Winterfeldt D, Fischer G W. Multi-attribute utility theory: models and assessment procedures[J]. Utility, probability, and human decision making, 1975: 47-85.

[17] Tzeng G H, Huang J J. Multiple attribute decision making: methods and applications[M]. CRC press, 2011.

[18] Zhu Y, Tian D, Yan F. Effectiveness of entropy weight method in decision-making[J]. Mathematical Problems in Engineering, 2020, 2020.

[19] Zhang Zhen, Huang Yanyan, Zhang Yongliang, Chen Tiande. Game confrontation algorithm of combat entity based on near end strategy optimization [J]. Journal of Nanjing University of technology, 2021,45 (01): 77-83

[20] Li Chen, Huang Yanyan, Zhang Yongliang, Chen Tiande. Multi agent decision making method based on actor critical framework and its application in wargame [J]. Systems engineering and electronic technology, 2021,43 (03): 755-762

[21] Liu man, Zhang Hongjun, Hao Wenning, Cheng Kai, Wang Jiayin. Intelligent decision making method of tactical wargame entity combat action [J]. Control and decision, 2020,35 (12): 2977-2985

[22] Dorton S L, Maryeski L A R, Ogren L, et al. A wargame-augmented knowledge elicitation method for the agile development of novel systems[J]. Systems, 2020, 8(3): 27.

[23] Zhang J, Xue Q, Chen Q, et al. Intelligent Battlefield Situation Comprehension Method Based On Deep Learning in Wargame[C]//2019 IEEE 1st International Conference on Civil Aviation Safety and Information Technology (ICCASIT). IEEE, 2019: 363-368.

[24] Vlachos I K, Sergiadis G D. Intuitionistic fuzzy information – applications to pattern recognition[J]. Pattern Recognition Letters, 2007, 28(2): 197-206.

[25] Kaelbling L P, Littman M L, Moore A W. Reinforcement learning: A survey[J]. Journal of artificial intelligence research, 1996, 4: 237-285.